# A Storage Expansion Planning Framework using Reinforcement Learning and Simulation-Based Optimization

Stamatis Tsianikas[a], Nooshin Yousefi[a], Jian Zhou[b], Mark D. Rodgers[c], and David W. Coit[a]

[a] Rutgers University, Department of Industrial & Systems Engineering, 96 Frelinghuysen Road, Piscataway, NJ 08854, USA
[b] Nanjing University of Science and Technology School of Economics and Management, Nanjing 210094, China
[c] Rutgers Business School Department of Supply Chain Management, 1 Washington Park Newark, NJ 07102, USA

**Abstract**

In the wake of the highly electrified future ahead of us, the role of energy storage is crucial wherever distributed generation is abundant, such as in microgrid settings. Given the variety of storage options that are becoming more and more economical, determining which type of storage technology to invest in, along with the appropriate timing and capacity becomes a critical research question. It is inevitable that these problems will continue to become increasingly relevant in the future and require strategic planning and holistic and modern frameworks in order to be solved. Reinforcement Learning algorithms have already proven to be successful in problems where sequential decision-making is inherent. In the operations planning area, these algorithms are already used but mostly in short-term problems with well-defined constraints. On the contrary, we expand and tailor these techniques to long-term planning by utilizing model-free algorithms combined with simulation-based models. A model and expansion plan have been developed to optimally determine microgrid designs as they evolve to dynamically react to changing conditions and to exploit energy storage capabilities. We show that it is possible to derive better engineering solutions that would point to the types of energy storage units which could be at the core of future microgrid applications. Another key finding is that the optimal storage capacity threshold for a system depends heavily on the price movements of the available storage units. By utilizing the proposed approaches, it is possible to model inherent problem uncertainties and optimize the whole streamline of sequential investment decision-making.

# 1. Introduction

## 1.1 Problem motivation

The functioning of modern society largely depends on continuous power supply. Electric power system interruptions can be disastrous, affecting hundreds of millions of people with inestimable costs (Campbell & Lowry, 2012; Henry & Ramirez-Marquez, 2016). We have already experienced massive cascading failures in electric power systems all over the world (J. Zhou, Huang, Coit, & Felder, 2018). Microgrids, which couple renewable technology with conventional technologies to meet stochastic energy demand, are anticipated to be increasingly significant in the future because of their advantages ranging from power resilience to renewable integration (Adam Hirsch, Yael Parag, & Josep Guerrero, 2018; Hu, Souza, Ferguson, & Wang, 2015). Efforts have been made by worldwide governments to stimulate renewable energy applications mainly due to environmental concerns (Guajardo, 2018). However, the unpredictable nature of some renewable energy sources results in intermittency regarding the power supply to microgrids (Aflaki & Netessine, 2017; Levron, Guerrero, & Beck, 2013) and therefore make the development of analytical and stochastic optimization methodologies to cope with them more needed than ever (Golari, Fan, & Jin, 2017). Effective energy storage can be helpful to alleviate this uncertain nature of renewable energy resources, such as wind power and solar power, by storing the energy at the time of low load and releasing the stored energy at the time of high load (Wu & Kapuscinski, 2013; J. Zhou, Tsianikas, Birnie, & Coit, 2019). Energy storage is accepted as an indispensable component of future microgrids, no matter if these work in stand-alone mode or not (Zhao et al., 2014). Moreover, it has recently become economically justifiable (Bahramirad, Reder, & Khodaei, 2012). Mallapragada et al. studied the problem of quantification of the long-run system value of battery energy storage in systems with significant and increasing wind and solar penetration (Mallapragada, Sepulveda, & Jenkins, 2020). It is therefore clear that the optimization of expansion planning in energy storage of the microgrids is crucial and affects millions of customers who currently, or will in the future, have their load demand served by microgrids.

## 1.2 Related work

Many mathematical optimization methods have been applied to solve energy storage expansion planning problem (Dehghan & Amjady, 2016; Hajipour, Bozorg, & Fotuhi-Firuzabad, 2015), such as linear programming, non-linear programming and mixed-integer liner programming, or heuristic optimization approaches, for example, genetic algorithm (Hemmati, Hooshmand, & Khodabakhshian, 2013). However, several real-world factors bring about more and more uncertainties into the problem, which makes it increasingly hard to find optimal expansion plans by using these traditional methods. Baringo et al. considered both short-term (demand variability, production of stochastic units etc.) and long-term (future peak demand, number of electric vehicles) uncertainties when they proposed an adaptive robust optimization approach for the expansion planning problem (Baringo, Boffino, & Oggioni, 2020). These uncertainties can though cover a much wider range of spectrum across the long-term planning problem. To illustrate this point, it is worthwhile highlighting two aspects: first, the problem of proper quantification of the value of lost load. In other words, the problem of accurately predicting the penalty cost for not meeting the energy demands of a given facility, at a given time and a given location. Secondly, another uncertainty brought into the area has to do with grid outages. Although the frequency and the duration of these can be measured and tracked, it is still a challenge to appropriately model them and create frameworks that take them into account. All the above could constitute some of the main drivers that render big data analytics and machine learning emerging research fields in operations management (Choi, Wallace, & Wang, 2018).

In this paper, Markov Decision Processes (MDP) and Reinforcement Learning (RL) algorithms are used to propose a novel approach for solving large-scale expansion planning problems by considering the stochastic and dynamic nature of the problem. The proposed dynamic algorithm answers all the critical questions, such as (1) whether it is actually necessary to add storage in the energy system, (2) when to install this storage, (3) how much capacity it should be added, and (4) which storage technology should be chosen. In contrast with the traditional optimization techniques that have been used in the past, our approach attempts to solve a stochastic and dynamic version of the problem, successfully capturing several inherent uncertainties. Moreover, it is equally important that the current work is able to provide a scalable methodology that could be utilized for real-scale microgrids.

Reinforcement Learning provides a mathematical framework for discovering or learning strategies that map situations onto actions with the goal of maximizing a cumulative reward function (R. S. Sutton & Barto, 1998). Reinforcement Learning has been used in various areas, such as transportation, maintenance, operation research, and energy systems. However, in the field of energy systems, Reinforcement Learning is used to solve mostly short-term planning problems, such as battery scheduling, unit commitment or building control. Several efforts also focused on the use case of hybrid electric vehicles (Han, He, Wu, Peng, & Li, 2019; Xiong, Cao, & Yu, 2018) . Dimeas and Hatziargyriou (Dimeas & Hatziargyriou, 2010) proposed a general framework of microgrids control based on a multi-agent Reinforcement Learning method. In a similar topic, Wang et al. conducted a review study on research works that attempt to utilize Reinforcement Learning-based approaches for building controls (Z. Wang & Hong, 2020). In another interesting work, Ebell et al. used an on-policy algorithm (SARSA) to tackle the problem of optimal power flow control in a PV+battery system (Ebell, Heinrich, Schlund, & Pruckner, 2018). What adds extra value to this work, is the fact that they focused their study to residential houses and use cases. Foruzan et al. (Foruzan, Soh, & Asgarpoor, 2018) used a multi-agent Reinforcement Learning framework to study distributed energy management in a microgrid considering the uncertainties involved in the nature of microgrids due to variability in renewable generation output power and continuous fluctuation of customers' consumption. In this work, there are two main interesting features, from the research perspective: first, how the authors design their agents/entities, with each of them (distributed energy resources, consumers and the main grid) having full control over their actions. Secondly, they use an interesting, and self-designed, array of evaluation metrics which could be inherited for similar works. At this point, it should be made clear that most of these research works study interesting microgrid-related problems, but however from the short-term scope; on the other side, we attempt to provide a novel framework, which combines Reinforcement Learning and Simulation, that can be exploited to tackle long-term planning problems in a more efficient way than traditional optimization techniques. This is clearly, and it was intended to be, one of the main novelties of the present work.

Zhou et al. (Y. Zhou, Scheller-Wolf, Secomandi, & Smith, 2019) used an MDP to develop a methodology that manages optimally a wind farm which is based on a storage-enabled grid-level facility. Rocchetta et al. (Rocchetta, Bellani, Compare, Zio, & Patelli, 2019) utilized Reinforcement Learning in conjunction with artificial neural networks, in order to tackle the problem of operation and maintenance planning in power grids. The authors formulated the stochastic environment under which the planning occurs and also defined strict constraints on the action sets. Li et al. (Li, Wu, He, & Chen, 2012) developed a Reinforcement Learning algorithm to minimize the electricity costs in a microgrid. Raju et al. (Raju, Sankar, & Milton, 2015) developed a multi-agent Reinforcement Learning algorithm for battery scheduling optimization in a microgrid. Lu et al. pooled several problem parameters together and used a Deep Reinforcement Learning approach to attack a slightly different problem: the one of optimal energy trading between neighboring microgrids (Lu et al., 2019). Rafique et al. focused on deriving dynamic models for determining the optimal and politically feasible ways to build up an energy supply chain under specific constraints (Rafique, Mun, & Zhao, 2017). It is intriguing how the authors used a macroeconomic perspective in their dynamic models and linked it with energy security and efficiency. This topic is particularly relevant here, because the present work can be considered as one of the first research attempts to combine Reinforcement Learning approaches with classic long-term problems of the energy sector. Kuznetsova et al. (Kuznetsova et al., 2013) used a two-step ahead Reinforcement Learning algorithm for battery scheduling in a microgrid. Leo et al. (Leo, Milton, & Sibi, 2014) proposed a three-step ahead Reinforcement Learning algorithm to optimize the battery scheduling in a dynamic environment. Mbuwir et al. (B. Mbuwir, F. Ruelens, F. Spiessens, & G. Deconinck, 2017) used a batch Reinforcement Learning to optimize operation scheduling of a storage device in a microgrid. Finally, Duan et al. studied the problem of optimal control for the specific case of hybrid energy storage systems, when combined with photovoltaic arrays and diesel generators (Duan et al., 2019).

We could summarize the main advantages of data-driven and Machine Learning-based approaches compared to "fit and forget" methods in two traits: agility and scalability. Concerning the former, it is clear that "compact" and static optimization approaches cannot easily adapt to the fast-paced industry of energy systems and the ever-changing characteristics of it. As to scalability, it is worthwhile to look at recent successful applications of Machine Learning and AI in various interesting problems of our era: many of them deal with a huge amount of (structured or unstructured) data. These successes are exactly what this current work attempts to exploit, for the benefit of microgrid planners and the energy industry in general. Table 1 summarizes all the similarities and differences of our proposed approach with the ones that can be found currently in the literature.

| Reference | Stochasticity | Detailed outage modeling | RL-based | Storage options other than batteries | Transmission planning | VOLL consideration | Long-term scope | Short-term scope |
|---|---|---|---|---|---|---|---|---|
| (Bahramirad et al., 2012) | ✔ | | | | | | ✔ | |
| (Hajipour et al., 2015) | ✔ | | ✔ | | | ✔ | ✔ | |
| (S. Q. Wang, Lu, Han, Ouyang, & Feng, 2020) | | | | | | | ✔ | ✔ |
| (Dehghan & Amjady, 2016) | ✔ | | | | ✔ | ✔ | ✔ | |
| (Duan et al., 2019) | ✔ | ✔ | ✔ | | | | | ✔ |
| (B. V. Mbuwir, F. Ruelens, F. Spiessens, & G. Deconinck, 2017) | ✔ | | ✔ | | | | | ✔ |
| (Ebell et al., 2018) | ✔ | | ✔ | | | | | ✔ |
| (Alsaidan, Khodaei, & Gao, 2018) | ✔ | | | | | ✔ | ✔ | |
| Our research | ✔ | ✔ | ✔ | | | ✔ | ✔ | |

Table 1 Key features of state-of-the-art storage planning models in the present literature

As elaborated in Table 1, in most of the previous studies, Reinforcement Learning is used to solve mostly short-term planning problems, while in this paper, a Reinforcement Learning algorithm is used to solve expansion planning problems on a multi-year horizon. Long-term perspectives and planning are becoming more and more important in the wake of technological advancements and governmental attempts to pursue ambitious goals for the future of renewables in the energy sector. To quote Dwight D. Eisenhower: "plans are useless, but planning is indispensable".

The paper is organized as follows. Section 1 provides the problem motivation, as well as the related work. Section 2 introduces the general framework of the problem such as microgrid formation, storage scheduling and investment scheme. Section 3 shows the mathematical formulation of the MDP and the Reinforcement Learning algorithm. A case study is used in Section 4 to show how the proposed method can find the optimal policy. Conclusions are drawn in Section 5.

The notation used in formulating the proposed model is listed as follows:

<div align="center">Nomenclature</div>

| | |
|---|---|
| $G$ | Set of existing facilities in microgrid |
| $VOLL^g$ | Value of lost load for a facility $g$, $/kWh |
| $C_p{}^g$ | Critical load factor for a facility $g$ |
| $SoC$ | State of charge of storage unit |
| $p_c{}^i$ | Charging proportion of storage unit $i$ |
| $p_d{}^i$ | Discharging proportion of storage unit $i$ |
| $B_r{}^i$ | Price of storage unit $i$, $/kWh |
| $Q^i(t)$ | Energy stored in storage unit $i$ at time $t$, kWh |
| $DoD^i$ | Depth of discharge of storage unit $i$ |
| $B^i{}_{min}$ | Minimum energy value of storage unit $i$, kWh |
| $e^i$ | Round-trip efficiency of storage unit $i$ |
| $P^i{}_{annuity}$ | Annual payment amount of storage investment for storage unit $i$, $ |
| $P^i{}_{principal}$ | Principal payment amount of storage investment for storage unit $i$, $ |
| $r$ | Annual discount or interest rate |
| $L^i$ | Lifetime of storage unit $i$, yrs |
| $C_k{}^{inv}$ | Investment cost for decision period $k$, $ |
| $C_k{}^{los}$ | Lost load cost for decision period $k$, $ |
| $P_{solar}(t)$ | Power production of solar array at time $t$, kWh |
| $P_{wind}(t)$ | Power production of wind turbine at time $t$, kWh |
| $\eta_{solar}$ | Efficiency of solar panel |
| $A_{cell}$ | Area of each solar cell, m$^2$ |
| $n_{cpp}$ | Number of solar cells per panel |
| $n_{pan}$ | Number of solar panels |
| $\eta_{wind}$ | Efficiency of wind turbine |
| $\rho$ | Air density, kg/m$^3$ |
| $A_{tur}$ | Area of wind turbine, m$^2$ |
| $I(t)$ | Solar irradiation at time $t$, kW/m$^2$ |
| $W(t)$ | Wind speed at time $t$, m/s |
| $K$ | Number of decision periods |
| $N_k$ | Set of outages for decision period $k$ |
| $O_{jk}$ | Set of time interval for outage $j$ for decision period $k$ |
| $W_{in}$ | Wind cut-in speed, m/s |
| $W_{out}$ | Wind cut-out speed, m/s |
| $CAIDI$ | Customer Average Interruption Duration Index, hrs/interruption |
| $SAIFI$ | System Average Interruption Frequency Index, interruptions/yr |
| $S^{tf}$ | Timing feature of state space |
| $S^{ef}$ | External feature of state space |
| $S^{if}$ | Internal feature of state space |
| $SU$ | Set of storage units in the system |
| $SC$ | Set of storage characteristics for each unit |
| $SL$ | Set of available expansion levels |
| $A_l$ | Amount of storage increase for expansion level $l$, kWh |
| $f^{tf}$ | Timing component of state transition function |
| $f^{ef}$ | External component of state transition function |
| $f^{if}$ | Internal component of state transition function |
| $p^{ef}$ | Transition matrix of the $s^{ef}$ DTMC |
| $r_k(s,a)$ | Reward function for the decision period $k$ of the problem given state $s$ and action $a$ |
| $n^{yrs}$ | Number of years in a decision period |
| $D(t,g)$ | Load demand for facility $g$ at time $t$, kWh |
| $\delta(t,g)$ | Indicator function for lost demand for facility $g$ at time $t$ |
| $f^{RF}$ | Random forest function for the outage cost component |

| $\alpha$ | Learning rate for Q-learning algorithm |
| $\gamma$ | Discount rate for Q-learning algorithm |
| $\varepsilon$ | Exploration/exploitation tradeoff parameter for Q-learning algorithm |

## 2. Problem framework

In this section, we present a detailed conceptual formulation of the problem under investigation. The primary objective of this analytical framework is to sequentially determine the optimal battery storage investment strategy to expand capacity for a system of distributed electricity generation plants connected in a microgrid network. It should be made clear at this point that there is only one scope at this problem: the long-term one. However, in order to efficiently estimate some cost components of the problem, we deploy a simulation-based technique, followed by a machine learning algorithm, that utilize operational level details (such as battery scheduling, outage modeling etc.). Therefore, although the MDP formulation in Section 3.1 does not depend on the short-term management of the microgrid, certain assumptions about the operational level should be clarified and afterwards used in the simulation-based approach in Section 3.2. Here, in the first subsection the microgrid formation is introduced. Secondly, the focus is given in the storage units, and more specifically in the assumptions behind their scheduling and investment. In the third subsection, renewable energy production and outage modeling are defined. Finally, the equations for the two cost components are given and the inherent trade-off between these two is explained.

### 2.1 Microgrid formation

In this problem formulation, we consider an interconnected system of microgrids, which is otherwise referred to as a family of community microgrids. Community microgrids are basically a natural extension of residential microgrids (A. Hirsch, Y. Parag, & J. Guerrero, 2018). These are again small-scale microgrids, which are comprised of various distributed electricity generation facilities, which must supply downstream customers with uninterrupted access to electricity at specified reliability (or customer service) levels. To mitigate the intermittency of generation output from these distributed generation facilities, these renewable plants are integrated with energy storage units to regulate their output in order to satisfy the demand of the multiple facilities in this network. However, in this context of this problem, these microgrids serve as reliable and uninterrupted backup generation resources, in the event that the centralized electric power grid experiences a disruption in service. This is particularly relevant and necessary for facilities that provide critical resources to a community that rely on its electricity supply, such as a hospital (Padilla, 2018). Each facility in the considered microgrid network is assigned its own value of lost load, $VOLL^g$, which is based on the financial damages associated with the inability to satisfy the demand at a given facility, and critical load factor $C_p^g$, which prioritizes the facilities in order of criticality (where $g \in G$ and $G$ is the set of existing facilities). These assumptions are crucial to the system design, since they affect how the energy produced by distributed plants, in tandem with their associated energy storage systems, is distributed to facilities in the network based on a prioritization scheme. Facilities are ranked based on their criticality and need to be served accordingly. Lastly, within the microgrid network, we consider a wide array of renewable energy plants, such as photovoltaic arrays, wind turbines and other distributed generation options to satisfy demand.

It is well-established in the literature that a mixture of different storage units, resulting in so-called Hybrid Energy Storage Systems (HESS), yield more beneficial microgrid network solutions (Faisal et al., 2018; Jing, Lai, Wong, & Wong, 2017). Therefore, a sufficient amount of storage options should be incorporated in the investment portfolio, and we thus aim to determine the optimal combination of storage technology investments to aid in supplying energy to the microgrid. Moreover, utilizing storage systems to augment distributed generation plants has been proven to yield benefits for the network, as they mitigate the intermittent generation output associated with renewable energy sources (Bocklisch, 2015; Tsianikas, Zhou, Yousefi, & Coit, 2019). However, one of the critical contributions of this work is that storage investments decisions are made in a sequential and dynamic fashion and can be revisited at various times within the planning horizon. This is in contrast to the existing models in the literature, where storage investment decisions are made at the beginning of each planning period and are not revisited in the future. This contribution enables system planners to account for various uncertain events in the planning horizon, such as the declining future projections in storage systems prices, in the investment planning process (IRENA, 2017).

## 2.2 *Storage operations and investment planning scheme*

A typical problem existing in all energy systems that contain different types of storage units is the charging and discharging scheduling of the storage systems. While it is often simplistically considered a standalone optimization problem, the biggest problem in the presence of multiple Energy Storage Systems (ESSs) is that a simultaneous discharging of paralleled ESSs would unavoidably result in significant state-of-charge (*SoC*) differences between the various storage units (Semënov, Mirzaeva, Townsend, & Goodwin, 2017). These differences, if propagated through several periods, could result in system power drops because some storage units would stop their operation earlier than others. This event may clearly put satisfying load demand at risk.

The potential solutions to this problem depend on whether the microgrid is controlled centrally or is decentralized. In the former case, there is a centralized control unit in the microgrid, which gathers all the necessary information, such as *SoC* levels and inherent storage unit characteristics and distributes the amount of energy provided by different ESSs in such a way that guarantees similar *SoC* levels among all the storage units in the system, while simultaneously satisfying the system load. On the contrary, in the latter case, there are various techniques that can be implemented, such as relating droop coefficients to the levels of *SoC* (Semënov et al., 2017; S. Q. Wang et al., 2020). By doing so, it is possible to force the higher-charged ESSs to contribute more active power than the lower-charged ones. The implementation used in this current work is more closely related to the decentralized approach mentioned above and is based on predetermined contribution ratios that are able to achieve the necessary *SoC* balancing, without utilizing *SoC* real-time information. Towards this direction, it is required to introduce the definitions of these charging and discharging ratios considered in this work. These ratios reflect the proportion of energy that each ESS should contribute while charging or discharging respectively. Therefore, the charging proportions $p_c^i$ and the discharging proportions $p_d^i$ are defined as follows:

$$p_c^i = \frac{B_r^i \; DoD^i}{e^i \sum\limits_{j \in SU} \dfrac{B_r^j \; DoD^j}{e^j}}, \forall i \in SU \tag{1}$$

$$p_d^i = \frac{B_r^i \; DoD^i \; e^i}{\sum\limits_{j \in SU} B_r^j \; DoD^j \; e^j}, \forall i \in SU \tag{2}$$

where $SU$ is the set of the various storage units existing in the microgrid, $B_r^i$ is the capacity of the $i^{th}$ storage unit, $DoD^i$ its corresponding depth-of-discharge and $e^i$ the round-trip efficiency. These parameters guarantee similar $SoC$ levels among the different storage units and also guarantee that:

$$\max_{i \in SU} SoC^i = 1 \Rightarrow SoC^i \approx 1, \forall i \in SU \tag{3}$$

$$\min_{i \in SU} SoC^i = 1 - DoD^i \Rightarrow SoC^i \approx 1 - DoD^i, \forall i \in SU \tag{4}$$

This technically means that all storage units reach simultaneously their maximum and minimum allowed levels of charge.

Concerning the monetary investment in storage units, an amortization model has been adopted where the payments are made annually, and each payment is calculated as follows:

$$P_{annuity}^i = P_{principal}^i \; \frac{r \left(1+r\right)^{L^i}}{\left(1+r\right)^{L^i} - 1} \quad , \forall i \in SU \tag{5}$$

where $P_{principal}^i$ is the principal investment amount of the $i^{th}$ storage unit, $r$ is the annual interest rate and $L^i$ is the lifetime of the $i^{th}$ storage unit. This amortization model resembles a leasing scheme, in which annual payments and the existence of the storage unit in the system are continued after the lifetime period of the unit expires. This may seem counterintuitive; however, it stems from the fact that storage units cannot be retired under the proposed approach. Although it would be ideal to include such decisions in the problem formulation, Equation (5) provides an easy way to incorporate information about the lifetime of the various storage options in the problem economics, without altering the information about their cost parameters.

*2.3    Renewable energy production and outage modeling*

Other assumptions are that solar and wind output power calculations has been adopted, based on solar cells per panel, available solar panels, rotor swept area of the wind turbine and the total number of wind turbines (Song, Li, Felder, Wang, & Coit, 2018). The equations are shown as follows:

$$P_{solar}\left(t_{ijk}\right) = \eta_{solar}\ A_{cell}\ n_{cpp}\ n_{pan}\ I\left(t_{ijk}\right),\ \forall i \in O_{jk},\ j \in N_k,\ k \in \{1,2,3,...,K\} \tag{6}$$

$$P_{wind}\left(t_{ijk}\right) = \begin{cases} 0, \text{if } W\left(t_{ijk}\right) \leq W_{in} \\ \dfrac{1}{2}\eta_{wind}\rho A_{tur}\ n_{tur}W\left(t_{ijk}\right), \text{if } W_{in} < W\left(t_{ijk}\right) < W_{out}, \\ 0, \text{if } W\left(t_{ijk}\right) \geq W_{out} \end{cases} \tag{7}$$

$$\forall i \in O_{jk},\ j \in N_k,\ k \in \{1,2,3,...,K\}$$

where $\eta_{solar}$ is the solar panel efficiency, $A_{cell}$ is the area of each solar cell, $n_{cpp}$ is the number of solar cells per panel, $n_{pan}$ is the number of solar panels in the system, $\eta_{wind}$ is the wind turbine efficiency, $\rho$ is the air density, $A_{tur}$ is the rotor swept area of the turbine and $n_{tur}$ is the total number of wind turbines in the system. Moreover, $I\left(t_{ijk}\right)$ and $W\left(t_{ijk}\right)$ denote the solar irradiance and the wind speed at time $t_{ijk}$ accordingly. $N_k$ is defined as a random set containing all grid outages that occurred in decision period $k$, while $O_{jk}$ is defined as a random set of all time intervals of grid outage $j$ in decision period $k$ and $K$ is the total number of decision periods. Finally, $W_{in}$ and $W_{out}$ are called cut-in and cut-out wind speeds and define the range in which the wind turbine can safely produce energy.

It should be further mentioned that a growth rate of 1% per year has been assumed for the load demand. For the outage modeling, the standard customer-oriented metrics Customer Average Interruption Duration Index (CAIDI) and System Average Interruption Frequency Index (SAIFI) are used in this framework as surrogates for customer service levels. CAIDI is reported as the average duration of an outage that a customer suffers in hours, while SAIFI is reported as the average interruptions on the system per year (Hyland, Doyle, & Yoon Lee, 2014).

*2.4    Cost-related components*

There are two main components that are considered in the cost function of this present research work: the first one is the investment cost in a specific storage unit and the second one is the cost of lost load when we are unable to meet the demand. Before proceeding with presenting the equations of these two costs, it is necessary to formally define how loss load is calculated. For this purpose, $\delta\left(t_{ijk}, g\right)$ is used as an indicator function on whether the load demand for facility $g$ at time $t_{ijk}$ is lost or not. The definition of the $\delta$ function, in which it should be assumed that the facilities in the set $G$ are ranked based on a prioritization scheme (i.e. facility 1 is the most critical, facility 2 is the second most critical, etc.) is as follows:

$$\delta(t_{ijk}, g) = \begin{cases} 1, & \text{if } Q^b\left(t_{ijk}\right) + \int_{t_{ijk}}^{t_{ijk}+\Delta t}\left(p_c^b\left(P_{solar}(u) + P_{wind}(u)\right) - \frac{p_d^b}{e^b}\sum_{m=1}^{g} C_p^m D(u,m)\right) du < B_{\min}^b \\ 0, & \text{otherwise} \end{cases}$$

$$\text{for any } b \in SU \text{ and } \forall i \in O_{jk}, j \in N_k, k \in K, g \in G \qquad (8)$$

In simpler terms, $\delta(t_{ijk}, g)$ is equal to 1 for the facility $g$ if the energy stored in storage units, defined as $Q^b$, combined with the energy production by renewable plants net the demand of facility $g$, defined in the second part of the left-hand side of the inequality in the first branch of Equation (8), are lower than the minimum allowed energy level $B_{\min}$. It should be noted here that $B_{\min}$ is determined by the depth-of-discharge of the storage unit. On the contrary, $\delta(t_{ijk}, g)$ is equal to 0, only when the system (energy in storage units and energy production by renewable plants net the demand of the facility $g$) is able to satisfy the demand for all the facilities up to $g$. Therefore, for the most critical facility ($g$=1), the system needs to be able to meet the demand only for this facility, in order for the load demand to be met. For the second most critical facility ($g$=2), the system needs to meet the demand for facility 1 and facility 2, etc. An important point is that just one arbitrary $b \in SU$ was chosen, in order to determine whether the demand is lost or not for a specific facility. The justification for this comes from Equation (4) which suggests that if one storage unit $b$ falls below the minimum allowed level $B_{\min}^b$, then the same should apply for the rest of the storage units. Finally, and because of the fact that the system is designed with the main purpose to be backup energy provider, the storage units are utilized only during grid outages. Therefore, it is clear that the larger the storage, the longer the system is able to satisfy the demand of facilities and the fewer are the times that the indicator function $\delta$ is equal to 1.

It is considered now appropriate to define the two aforementioned cost components for the $k^{\text{th}}$ decision period:

$$C_k^{inv} = \sum_{m \in SU} \sum_{l \in SL} \left(K - k + 1\right) n^{yrs} a_{m,l} \, P_{annuity}^m \tag{9}$$

$$C_k^{los} = \sum_{g \in G} VOLL^g \sum_{j \in N_k} \sum_{i \in O_{jk}} \delta\left(t_{ijk}, g\right) C_p^g \, D\left(t_{ijk}, g\right) \tag{10}$$

Equation (9) presents the investment component, which is calculated as $\left(K - k + 1\right) n^{yrs}$ equal payments of $P_{annuity}^m$ for each storage unit $m$, where $n^{yrs}$ denotes the number of years in one decision period and $\left(K - k + 1\right)$ is the number of remaining decision periods in the problem horizon. By utilizing the definition of the indicator function $\delta(t_{ijk}, g)$, Equation (10) defines the loss of load cost, for all $G$ facilities in the formulation. $D\left(t_{ijk}, g\right)$ defines the load demand for facility $g$ at time $t_{ijk}$. Between these two cost components, there is an inherent trade-off from the perspective of the microgrid planner: if the overall strategy is to invest only minimally in storage units, the investment cost component should stay at the lowest range, but the loss of load cost is going to skyrocket. On the contrary, if the strategy is to invest heavily in storage units, the opposite behavior should be observed: the investment cost should become much higher, but the loss of load cost is going to be minimized. Finding the sweet spot between these two extreme strategies, along with the correct timing and appropriate selection of storage type, is the bulk of this research work.

## 3. MDP formulation of the expansion planning problem

Learning from interaction and achieving a goal is the main and sole purpose of Reinforcement Learning. This process is often described by a specific class of stochastic processes known as MDPs. Key elements of MDPs are the notions of the agent and the environment (R. Sutton & Barto, 2015). The agent acts as the decision-maker in the problem and the one who is responsible for learning. The environment is the main entity that the agent communicates with to obtain information. The agent and the environment continuously interact with one another in a process described as follows: the agent takes actions and the environment, and based on these actions, gives feedback to the agent called a reward. Overall, the agent aims to maximize its total earned rewards over a finite (or infinite) time horizon. This process is illustrated in Fig. 1:

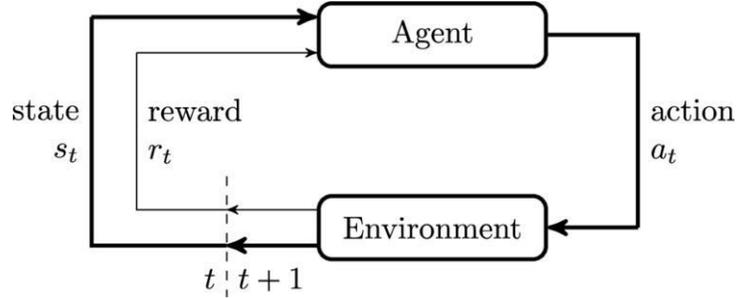

Fig. 1 Agent-environment interactions in Reinforcement Learning setting (R. Sutton & Barto, 2015)

To be more specific, the agent and the environment interact at specific discrete time steps, $t = 0, 1, 2, 3...$ At each time step $t$, the agent receives a representative description of the environment's state $S_t \in S$, where $S$ is the set of possible states of the environment and selects an action $A_t \in A(S_t)$, where $A(S_t)$ is the set of possible actions in state $S_t$. Consecutively, the environment sends back a numerical signal to the agent, which is a function of the agent's chosen action. This signal is called a reward in this context and is denoted $R_{t+1} \in R$. The purpose of the agent is to maximize the discounted sum of rewards as defined by:

$$G_t = \sum_{k=0}^{K-t-1} \gamma^k \, R_{t+k+1} \tag{11}$$

where $\gamma$ denotes the discount rate and $K$ is the problem horizon.

The agent then is responsible to do a mapping at each time step from states to actions. This mapping is called the agent's policy, denoted by $\pi_t$, where $\pi_t(\alpha|s)$ refers to the probability that $A_t = \alpha$, given that $S_t = s$. Finally, the system transitions to a new state $S_{t+1}$ and this procedure continues iteratively until convergence is reached (R. Sutton & Barto, 2015). Lastly, the value of taking action $a$ while in state $s$ and following policy $\pi$ can be formally defined as:

$$q_\pi\left(s,a\right) = E_\pi\left[G_t \mid S_t = s, A_t = a\right] = E_\pi\left[\sum_{k=0}^{\infty} \gamma^k R_{t+k+1} \mid S_t = s, A_t = a\right] \tag{12}$$

in which $q_\pi$ is intuitively called the action-value function of policy $\pi$ or otherwise the q-value function. This function is utilized to derive the optimal policies for the existing environment.

The solver algorithm for this problem is a variant of the well-known Q-learning algorithm (Watkins & Dayan, 1992). Q-learning is a model-free, off-policy learning algorithm which uses the following update rule for its q-values:

$$q\left(s,a\right) \leftarrow q\left(s,a\right) + \alpha\left[r + \gamma \max_a q\left(s',a\right) - q\left(s,a\right)\right] \tag{13}$$

where $\alpha$ here denoting the learning rate of the algorithm, i.e. how fast to approach the optimal solution. Equation (13) is derived from the famous Bellman equation, which is a fundamental dynamic programming mathematical equation.

Despite its simplicity and its high utilization in many MDP settings, Q-learning suffers from serious bias issues. It has been shown that the algorithm may have very poor performance in stochastic MDPs due to large overestimation of action values (van Hasselt, 2010). This overestimation stems from the fact that positive bias is inherent to the Q-learning algorithm as a result of using the maximum action value as an approximation of the maximum expected action value. Q-learning uses the single estimator approach for estimating the value of the next state; $\max_a q\left(s',a\right)$ is an estimate for $E\left[\max_a q\left(s',a\right)\right]$, but then, in turn, it is used as an estimator for $\max_a E\left[q\left(s',a\right)\right]$.

To summarize, in the first subsection, we redefine the long-term dynamic storage investment problem in the context of the MDP framework, which is the fundamental basis of the solution algorithm. Specifically, a detailed definition of the state and action sets and the reward and transition functions are provided. In this research work, a novel approach is proposed to mitigate the overestimation bias problem of the Q-learning, by using synthetic datasets and metamodeling. Therefore, in the second subsection, the details and assumptions of the aforementioned approach are presented. Lastly, and before proceeding with case study and results, the final step-by-step algorithm is presented.

### 3.1 Markov Decision Process formulation

Every MDP, as defined above, is a tuple of $(S, A, f, R, \gamma)$, thus implying that it is fully defined with the state and action sets $S, A$, the transition and reward functions $f, R$ and the discount factor $\gamma$. Therefore, it is considered necessary to provide the required definitions for these elements, in order to be able to use the appropriate algorithms to derive optimal policies.

Starting with the state space $S$ of the problem, it should be mentioned that it consists of three sub-features; time features, external features, and internal features:

$$
\begin{aligned}
S &= S^{tf} \times S^{ef} \times S^{if} \\
\text{where:} \quad & s^{tf} \in S^{tf} = \left\{1, 2, ..., K\right\} \\
& \mathbf{s}^{ef} = \left(s_{i,j}^{ef}, \forall i \in SU, j \in SC\right) \in S^{ef} \\
& \mathbf{s}^{if} = \left(s_i^{if}, \forall i \in SU\right) \in S^{if}, \ \ \forall i \in SU, s^{tf} \in S^{tf}
\end{aligned}
\tag{14}
$$

$S^{tf}$ is the time-dependent component of the state space, which simply denotes the current decision period. It should be noted here that it is highly advised for the timing feature to be explicitly included in the state information of the problem. It has been proven that the agent's learning performance is significantly improved when time-awareness of the agent is introduced, by specifically incorporating a time-related space component (Pardo, Takavoli, Levdik, & Kormushev, 2018) (Harada, 1997). $S^{ef}$ defines the set of external features of the problem, such as the price, the efficiency and the depth-of-discharge of the storage unit, where $SU$ is again the set of storage units and $SC$ is the set of storage characteristics included in the formulation. They are called external because the information coming from these characteristics is received from the environment with no ability of the agent to affect them. For instance, a realization of $S^{ef}$ is the vector $\mathbf{s}^{ef}$, consisting of all the elements $s_{i,j}^{ef}$ which denotes the value of the $j^{\text{th}}$ characteristic of the $i^{\text{th}}$ storage unit. Finally, $S^{if}$ is the set of internal features of the problem, such as the storage capacity already installed in the system. They are called internal because the agent is able to affect this component by taking appropriate actions. Therefore, the microgrid's state is defined by a vector such as $\mathbf{s} = \left( s^{tf}, \mathbf{s}^{ef}, \mathbf{s}^{if} \right)$.

Concerning the action set of the problem, it is clearly defined based on the possible actions that the agent can take. In the context of the current problem, the agent should choose between taking no action or deciding to expand the storage capacity of a specific storage unit. If the latter is the case, the agent should do so at one of the available predetermined levels, in order to align with the discrete time and space assumptions of a Discrete Time Markov Chain (DTMC) framework. Therefore, the agent's action can be defined in vector form as follows:

$$\begin{aligned} &\boldsymbol{\alpha} = \left( a_{i,l}, \forall i \in SU, l \in SL \right) \in A \\ &\text{s.t.} \quad \sum_{i \in SU} \sum_{l \in SL} a_{i,l} \le 1 \\ &\qquad a_{i,l} \in \{0,1\}, \forall i \in SU, j \in SL \end{aligned} \qquad (15)$$

where $a_{i,l}$ denotes the binary action of expanding the capacity of the $i^{th}$ storage unit at the $l^{th}$ level, and $SL$ is the set of available expansion levels. The first constraint imposed in (15) guarantees the binarity of the action components, while the second constraint limits the agent, so it cannot take more than one expansion actions per period.

Proceeding with more definitions, the focus is now given to the state transition function $f$. Given the fact that the state is composed by three components (time, external and internal component), and by using the notation $s$ for the current state and $s'$ for the next state, the state transition equations are provided below:

$$s^{tf\,\prime} = f^{tf}\left( s^{tf} \right) = s^{tf} + 1, \quad \forall s^{tf} \in S^{tf} \qquad (16)$$

$$\mathbf{s}^{ef\,\prime} = f^{ef}\left( \mathbf{s}^{ef} \right), \text{ where: } \left\{ s_{i,j}^{ef}, s^{tf} \in S^{tf} \right\} \text{ is a DTMC with } p_{i,j}^{ef}, \forall i \in SU, j \in SC \qquad (17)$$

$$\mathbf{s}^{if\,\prime} = f^{if}\left( \mathbf{s}^{if}, \boldsymbol{\alpha} \right), \text{ where: } s_i^{if\,\prime} = s_i^{if} + \sum_{l \in SL} A_l\, a_{i,l}, \forall i \in SU \qquad (18)$$

Equation (16) is the state transition equation for the time feature of the state space and is simply an incremental by-one operation. Equation (17) preserves the Markov property of the external features of the state space; it means that the $j^{\text{th}}$ characteristic of the $i^{\text{th}}$ storage unit follows a DTMC with the corresponding $p_{i,j}^{ef}$ transition matrix. Equation (18) is the transition equation for the internal feature of the state space. Thus, if the decision has been made to expand the $i^{\text{th}}$ unit's storage capacity at the $l^{\text{th}}$ level, the corresponding $s_i^{if}$ is going to be increased by $A_l$ accordingly. Subsequently, the next state could be described by the vector $\mathbf{s}' = \left( f^{tf}\left( s^{tf} \right), f^{ef}\left( \mathbf{s}^{ef} \right), f^{if}\left( \mathbf{s}^{if}, \boldsymbol{\alpha} \right) \right)$.

The last component of the MDP to be defined in this context is the reward function. This is a crucial element of this definition, since it affects how the agent receives signals (i.e., rewards) from the environment. These signals are the main drivers that guide the agent to the derivation of the optimal policies. Therefore, the reward received at the $k^{\text{th}}$ decision period can be defined as:

$$r_k\left( \mathbf{s}, \boldsymbol{\alpha} \right) = -\sum_{m \in SU} \sum_{l \in SL} \left( K - k + 1 \right) n^{yrs} a_{m,l} \, P_{annuity}^m - \sum_{g \in G} VOLL^g \sum_{j \in N_k} \sum_{i \in O_{jk}} \delta\left( t_{ijk}, g \right) C_p^g \, D\left( t_{ijk}, g \right) \qquad (19)$$

Equation (19) is simply the addition of the two cost components presented in Equations (9) and (10). However, there is now a negative sign, due to the fact that the goal of the agent is to maximize its accumulated rewards.

### 3.2    *Utilization of synthetic datasets to tackle overestimation bias*

In the introduction of Section 3 it was shown why the Q-learning algorithm suffers from overestimation bias in highly stochastic environments. Practically, this means that if the agent assumes that there is a chance it would receive an extremely "good" reward moving to specific state, he may try to transition to that state, even though the optimal strategy would be to transition to other states. In this subsection of this paper, it is explained how this phenomenon applies to the examined case and a way that could potentially mitigate this effect is proposed. It should be mentioned that this is the only part of the current research work that the operational level assumptions defined in Section 2 are going to be used. The reason is that we want to construct a computationally efficient and unbiased way to estimate the loss of load cost component of the reward function in Equation (9).

It can be safely assumed that the problem arises from situations where the agent may obtain misleading "signals" on how the optimal strategy is structured. In this context, these signals correspond to the rewards that the agent receives in every decision period of the problem. The reward function, defined in Equation (19), is mainly composed of two negative components; the investment cost and the outage penalty. While the investment cost is clearly affected solely by the decision to expand storage capabilities, the outage penalty relies heavily on the stochastic events of outages. Considering the scenario of having 0 (or at least very few and/or short-lived) outages in a specific decision period, the agent may consider it beneficial for the system to proceed "as-is" and endure the outages, instead of taking actions to mitigate against them, i.e., investment actions. This result leads to the misleading "signals" that were previously mentioned. In the most favorable scenario, this phenomenon would ultimately slow down the convergence rate of the solution algorithm. The least favorable scenario could result in deriving sub-optimal policies. Consequently, it necessary to design an original approach for mitigating this effect using synthetic datasets and function approximation for the outage cost component of the reward function to address this challenge.

As its name suggests, synthetic datasets consist of data observations that are generated programmatically via simulation techniques, and not by real-life experiments and data collection (KDnuggets, 2018). In this case, simulation techniques such as those outlined in (J. Zhou et al., 2019) and (Tsianikas, Zhou, Birnie, & Coit, 2019) can be utilized in order to generate a synthetic dataset consisting of multiple input features and a single output feature, i.e., the outage cost. Subsequently, a function approximation technique can be used to map the given inputs to the desired, and approximate, output. Therefore, the features needed to predict the outage cost can create a vector of the following form: $\left( s^{tf}, s_i^{if} \right) \forall i \in SU$, meaning that this specific cost component depends on the timing feature of the state space and the installed capacity of every storage unit in the system. In this context, $s_i^{if}$ denotes the installed capacity of the $i^{\text{th}}$ storage unit.

As mentioned in the first step of this process, there is a need to derive a systematic method to generate observations to be added to the synthetic dataset. Each of these observations comes from running $n$ individual and independent simulation runs of the system and averaging the obtained results. In order to generate independent observations, a random sample of the input features can be used, which implies that each input feature of the dataset (timing feature and installed capacities for all storage units) is arbitrarily selected from specified corresponding ranges. Moreover, for each of these individual simulation runs, outages are generated using the standard reliability (or customer service) metrics of CAIDI and SAIFI. More specifically, the duration of a specific outage is described by a shifted Poisson distributed random variable with mean CAIDI, while the outage events form a Poisson process with rate SAIFI. After the input features are selected for a specific observation and the outages are generated for each trial, then we generate $n$ simulation realizations of the system and the corresponding outputs (outage cost) are determined by averaging the results of these $n$ simulations. This procedure is iteratively followed until an $S$-sized dataset is created, where $S$ is the predetermined desired length of the dataset. Lastly, the random forest algorithm is used as a function approximation for the outage cost, given the synthetic dataset. Therefore, it is now seen that Equation (19) can be rewritten such as:

$$r_k\left(\mathbf{s}, \boldsymbol{\alpha}\right) = -\sum_{i \in SU} \sum_{l \in SL} \left(K - k + 1\right) n^{yrs} a_{i,l} P_{annuity}^i - f^{RF}\left(k, s_i^{if}\right) \forall i \in SU \tag{20}$$

where the second part (outage cost) is now the result of the function approximation technique (random forest) that was selected.

### 3.3    Final algorithm

Before proceeding with the numerical case studies and results, we provide a schematic and holistic representation of the proposed approach. Although the basis of the algorithm used is still the classing Q-learning approach, the preprocessing step of synthetic data creation and function approximation was added. The procedure can be seen in Table 2.

| **Algorithm:** Q-learning with preprocessing step |
|---|
| 1:    **initialization:** random outages for all simulation runs |
| 2:    **for** *every observation* **do:** |
| 3:            select arbitrarily $\left(k, \left(s_i^{if}, \forall i \in SU\right)\right)$ |
| 4:            **for** *every simulation run* **do:** |
| 5:                    simulate system and compute outage cost |

```
6:          end for
7:          average over all runs and store observation in the synthetic dataset
8:      end for
9:      use random forest to derive $f^{RF}$ from the synthetic dataset
10:     initialization: Q table
11:     for every episode do:
12:         initialization: starting state $s$
13:         for every decision period do:
14:             select $\alpha$ based on Q and $\varepsilon$-greedy policy
15:             observe $r\square\left(a,f^{RF}\right)$ and $s'$ transition function $f$
16:             $Q(s,a)\leftarrow Q(s,a)+\alpha\left[r+\gamma\max_{a}Q(s',a)-Q(s,a)\right]$
17:             $s\leftarrow s'$
18:         end for
19:     end for
```

Table 2 Schematical representation of the Q-learning algorithm with preprocessing step

The first nine lines of the algorithm define the preprocessing step and the last ten lines compose the typical steps of the Q-learning algorithm, adjusted for the current problem. The main reason for the mitigation of the overestimation bias problem that the Q-learning algorithm imposes comes from the seventh line of the proposed approach; the fact that the average over a large number of simulation runs is used and then a random forest algorithm is trained based on the synthetic dataset in order to get an estimation of the outage cost, makes the "signal" that the agent perceives much clearer and without unnecessary variance.

## 4. Case study and discussion

A case study is conducted using the methodologies described in this paper. The microgrid considered in this case study consists of several facilities (hospitals, schools, and residential houses). The selection of such a critical facility as a hospital dictates that the primary role of the designed energy storage system should be to provide backup energy during outages of the main grid. Hospitals are nowadays using electronic equipment such as operating room machinery, life support, blood storage etc. Therefore, they require uninterrupted operation with almost no exceptions at all (Tsianikas, Zhou, Birnie, et al., 2019). For the above reason, discharging of the storage units during normal grid operation is not allowed and the storage units are considered fully charged before every (if any) grid outage. The microgrid is located in Westhampton, NY. In Fig. 2, the whole area of Westhampton can be seen as a satellite view:

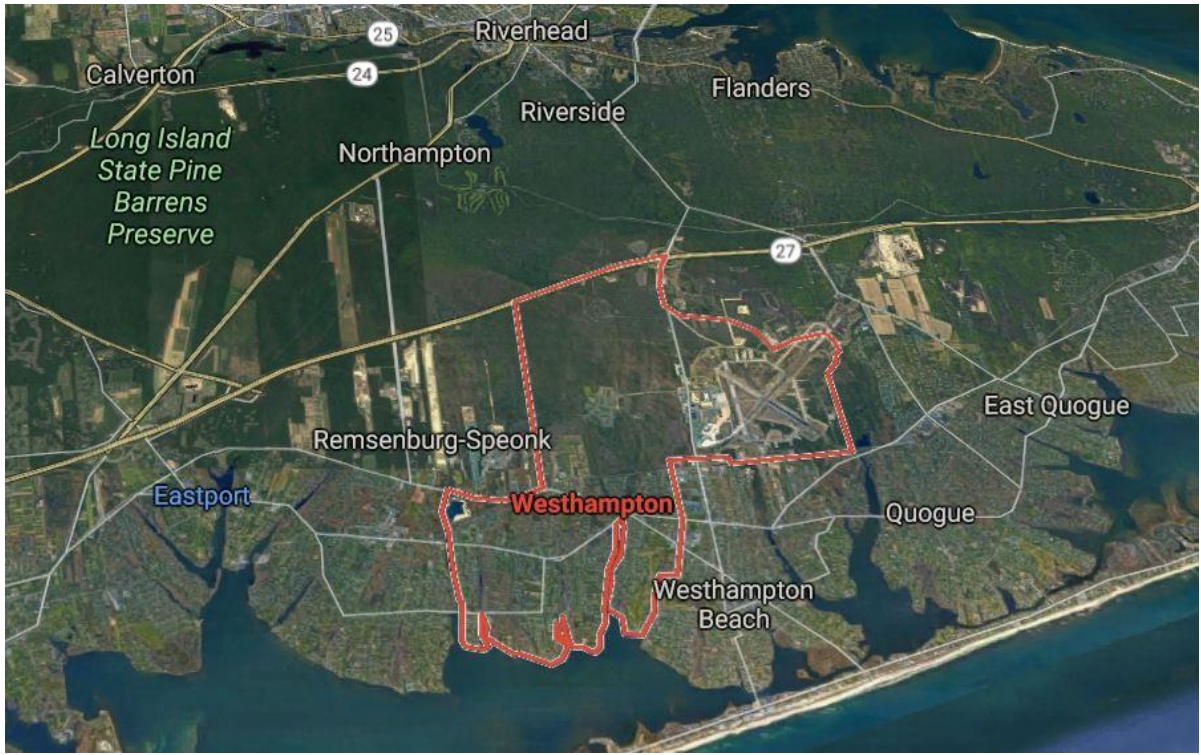
Fig. 2 Satellite view of the Westhampton, NY area (Google, 2019)

The reason that this particular location is chosen lies in the fact that this is an area with a high wind energy potential, very close to the North Atlantic Ocean. Location-specific demand and meteorological data are used and can be found in (NREL, 2013) (NREL, 2016). Each facility in the microgrid is assigned a *VOLL* and critical load factor. Concerning the storage options existing in the formulation, four different types of storage technologies are considered: Li-ion battery, lead-acid battery, vanadium redox battery, and flywheel storage system. These options were selected in order to explore various storage options, including not only the industry standard electrochemical storage systems, but other less common alternatives. Each storage type has its own characteristics, which of course are expected to affect the results in a significant fashion. It had been assumed that all the storage system characteristics can be described by a different deterministic function of the decision period, except the storage system price which holds its stochastic nature. The stochasticity of the storage price is modeled using Markov Chains, as described in the previous section. To further illustrate this concept, the Markov Chain used to model the storage price of the first storage unit (Li-ion) is given in Fig. 3.

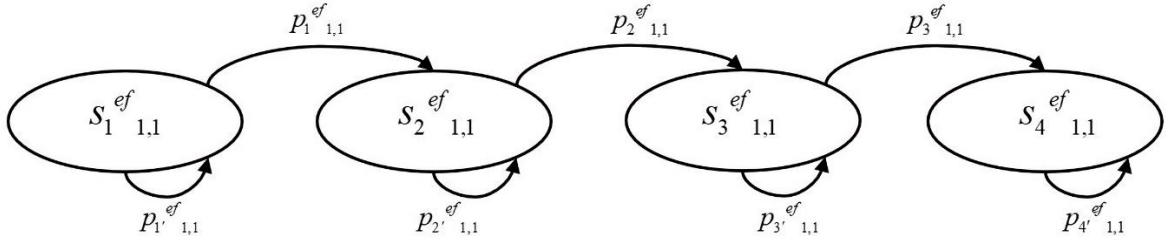

Fig. 3 Markov Chain for the price of Li-ion storage type

It should be noted here that the indices $i = 1$, $j = 1$ are used to refer to the 1st storage unit and its 1st characteristic respectively.

### 4.1    Numerical assumptions

Regarding the numerical assumptions of the case study, a 20-year time horizon was considered, where the decisions are made every 5 years, thus yielding 4 decision periods in total. The interest rate for storage investment is considered 2% annually. Solar and wind power plants are considered in the microgrid and their production was calculated using Equations (6) and (7) and by considering (Song et al., 2018) (J. Zhou et al., 2019): $\eta_{solar} = 0.16$, $A_{cell} = 0.0232258 \, m^2$, $n_{cpp} = 72$, $n_{pan} = 6000$, $W_{in} = 3 \, m^2$, $W_{out} = 22 \, m^2$, $\eta_{wind} = 0.48$, $\rho = 1.25 \, ^{kg}/_{m^3}$, $A_{tur} = 1520.53 \, m^2$, $n_{tur} = 10$. CAIDI and SAIFI are considered 5.122 and 1.155 respectively (Service, 2018) (Service, 2017). According to the facilities, there are three different types: hospital, school, and residential houses. The assumptions made for these facilities can be seen in Table 3 (van der Welle & van der Zwaan, 2007) (Alsaidan et al., 2018).

| Data  Facility | Number | VOLL | $C_p$ |
|---|---|---|---|
| Hospital | 2 | 25 | 0.8 |
| School | 5 | 17 | 0.6 |
| Residential | 300 | 8 | 0.4 |

Table 3 VOLL, critical load factor and count for the considered facilities

In the context of this problem, the agent has the option to choose from three discrete capacity levels for each storage unit and for each decision period. However, the agent is restricted to choose one action at maximum for each decision period, according to Equation (15). The storage capacity levels used in this case study are 300, 1000 and 3000 kWh. The various storage systems characteristics for each decision period of the problem can be seen in Tables 4-7 (IRENA, 2017). IRENA provides the current parameters, as well as forecasts for the year 2030. Using simple extrapolation/interpolation techniques, we can get values pertinent to the time horizon of the present work.

| Period / Li-ion | 1 | 2 | 3 | 4 |
|---|---|---|---|---|
| State for price MC ($/kWh) | 420 | 310 | 167 | 150 |
| Probability $p_{t,Li}^q$ for price MC | 0.70 | 0.70 | 0.70 | 0 |
| Lifetime (yrs) | 12 | 17 | 19 | 20 |
| Efficiency | 0.95 | 0.96 | 0.97 | 0.98 |
| DoD | 0.90 | 0.90 | 0.90 | 0.90 |

Table 4 Li-ion characteristics for all decision periods

| Period / Lead-acid | 1 | 2 | 3 | 4 |
|---|---|---|---|---|
| State for price MC ($/kWh) | 142 | 115 | 77 | 65 |
| Probability $p_{t,Li}^q$ for price MC | 0.70 | 0.70 | 0.70 | 0 |
| Lifetime (yrs) | 9 | 11 | 13 | 14 |
| Efficiency | 0.80 | 0.81 | 0.83 | 0.84 |
| DoD | 0.55 | 0.55 | 0.55 | 0.55 |

Table 5 Lead-acid characteristics for all decision periods

| Period / Vanadium redox | 1 | 2 | 3 | 4 |
|---|---|---|---|---|
| State for price MC ($/kWh) | 385 | 255 | 120 | 95 |
| Probability $p_{t,Li}^q$ for price MC | 0.70 | 0.70 | 0.70 | 0 |
| Lifetime (yrs) | 13 | 17 | 20 | 21 |
| Efficiency | 0.70 | 0.73 | 0.78 | 0.79 |
| DoD | 1 | 1 | 1 | 1 |

Table 6 Vanadium redox characteristics for all decision periods

| Period / Flywheel storage | 1 | 2 | 3 | 4 |
|---|---|---|---|---|
| State for price MC ($/kWh) | 3100 | 2600 | 1950 | 1700 |
| Probability $p_{t,Li}^q$ for price MC | 0.70 | 0.70 | 0.70 | 0 |
| Lifetime (yrs) | 20 | 26 | 30 | 32 |
| Efficiency | 0.84 | 0.85 | 0.87 | 0.88 |
| DoD | 0.86 | 0.86 | 0.86 | 0.86 |

Table 7 Flywheel storage characteristics for all decision periods

Finally, we present the algorithmic assumptions considered in this case study. Firstly, regarding the simulated synthetic data collection, a total of 1000 observations were generated, where each observation was produced after 100 simulation trials on the system. Herein, it should be mentioned that the accuracy of the simulated outages is verified using the same number of simulation trials and compared with the existing real data of CAIDI and SAIFI, which are 5.122 and 1.155. The corresponding synthetic numbers are very close and equal to 5.16 and 1.21 respectively. For the random forest model that was used to approximate the cost component related to outages, the dataset is split to train/test using a 0.8/0.2 ratio and a total number of 10 forests is used. Lastly, with respect to the Q-learning algorithm, a total of $10^7$ number of episodes is used, assuming $\gamma = 0.9$ with linearly decaying rates $\alpha$ and $\varepsilon$ ranging from 1 to 0.02. The decision for a linearly decaying exploration/exploitation tradeoff parameter $\varepsilon$ is very important, as it dictates the performance of the algorithm (Dearden, Friedman, & Russell, 1998). It means that it would be ideal to explore as much as possible at the initial episodes, while it would be better to approach convergence and just exploit the acquired knowledge at the final episodes.

### 4.2    Results and discussion

As a first step in presenting the results of the case study, we examine the performance of the random forest model for approximating the outage cost. The model obtained an R-squared score of 0.98 on the test set, which implies that the model used explains the variability in the dependent variable very well. Theoretically, the outage costs are likely to follow a decreasing function of the capacity already installed in the system. However, a decaying rate for this behavior is expected, meaning that the benefit associated with adding more capacity of a specific storage type in the system is negligible after a point where the already installed capacity is large enough. Given also the distribution of the outage durations, we can expect to see an initial range in the horizontal axis where adding more capacity does not result in significant outage cost savings. The aforementioned features can be observed in Fig. 4 for the storage units of Li-ion and vanadium redox:

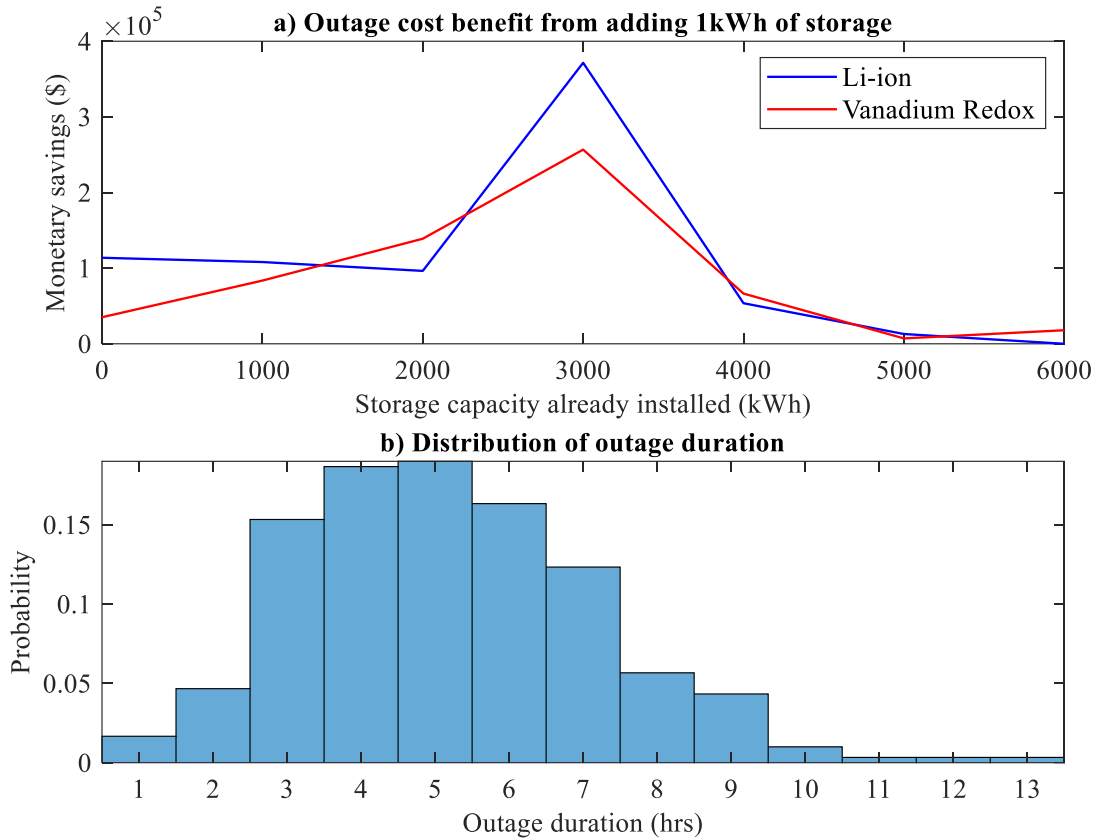

Fig. 4 a). Outage cost savings b). Distribution of outage duration

Before proceeding with analyzing Fig. 4, it should be made clear that Fig. 4 corresponds to the initial state of the environment and does not imply any overall superiority of the Li-ion battery over the vanadium redox one. It is clear from observing Fig. 4a) that the expected results were obtained. The behavior is similar for both storage types; after the initial phase where the cost savings for adding capacity are small, a peak is observed where the installed capacity is around 3000kWh. After that point, the cost savings are still positive, but approach zero. It is also very interesting to examine Fig. 4a) in accordance with Fig. 4b). To recall, the duration for each outage simulated follows a shifted Poisson distribution with mean approximately equal to 5.122. Fig. 4b) presents the approximate probability that a random outage obtains a value in the range of the horizontal axis. There is a clear threshold around 7 hours where afterwards, the outage events dramatically reduce in frequency. This implies, that when capacity levels are sufficient enough to mitigate against a large number of outages, it no longer becomes cost-effective for the planners to expand storage capacity in the system.

After verifying that the performance of the random forest model, we next observe the optimal policies derived from the proposed methodology. To recall, in order to extract optimal policies from the results, the output of the Q-learning algorithm is a completed table with each field denoting the q-value of each state-action pair. The amount of knowledge that the Q-learning is able to produce strongly depends on the number of episodes that the agent experiences. In the case study, each state of the environment is actually a tuple of 9 elements: the first element was the timing feature, the next 4 elements were the price states for each storage technology and the last 4 elements were the installed capacity again for each storage unit. The state of the environment is visually represented in Fig. 5:

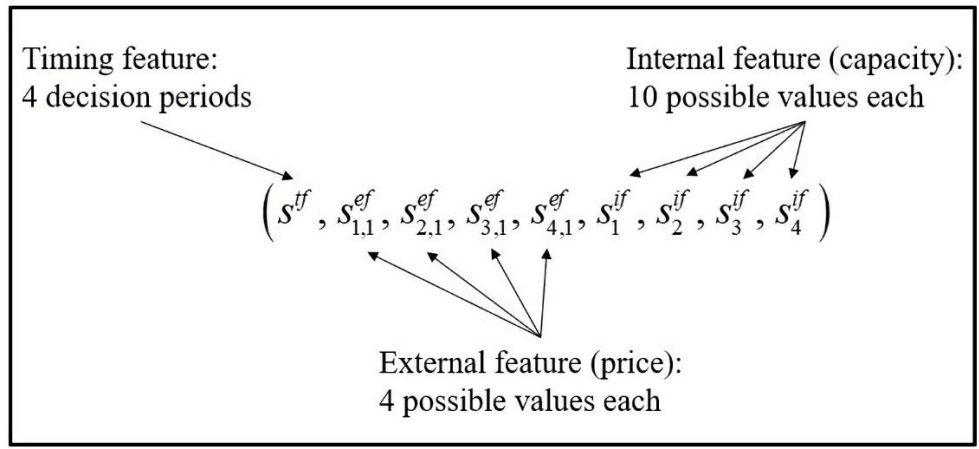

Fig. 5 Visual representation of the state of the environment for our case study

Therefore, the total number of states in the system can be calculated to be 2,758,578 states. This derivation comes from the fact that the state for the 1$^{st}$ decision period is fixed; for the 2$^{nd}$ period, there are 2 possible values for the external feature (price) of each storage unit and 4 possible values for the internal feature (capacity) of each storage unit; for the 3$^{rd}$ period, there are 3 possible values for the external feature (price) of each storage unit and 7 possible values for the internal feature (capacity) of each storage unit; finally, for the last period there are 4 possible values for the external feature (price) of each storage unit and 10 possible values for the internal feature (capacity) of each storage unit. If we also multiply by the number of possible actions in each decision period, the outcome is a total of 35,861,514 state-action pairs. Given the large magnitude of the state-action space, the only feasible way for observing the results of the proposed approach is to derive scenarios for price movement in the MDP and obtain optimal policies for each scenario separately.

Towards this direction, the first 3 scenarios are defined and examined. Scenario 1 refers to the case where the price of each storage unit is declining in every time period. Referring to Fig. 3, this scenario corresponds to the case where all the forward transitions are realized. Scenario 2 describes the case where again all storage prices are declining, except the price of vanadium redox unit for the periods 1, 3 (it only declines in period 2). Finally, scenario 3 refers to the case where all storage prices are again declining, except the price of Li-ion battery for periods 1, 3 (it only declines in period 2). Results can be observed schematically in Fig. 6:

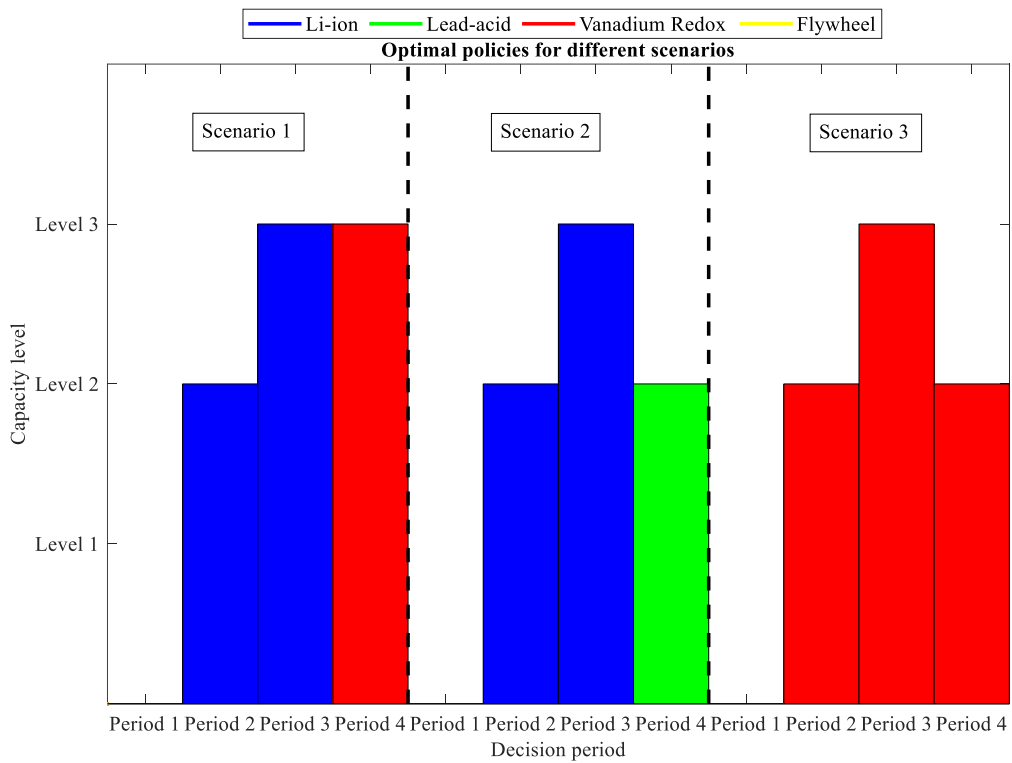

Fig. 6 Optimal policies derived for specific scenarios

The results in Fig. 6 reveal some very interesting trends. The "baseline" scenario 1 presents the optimal policy under which nothing should be done in the first decision period, Li-ion battery should be installed at level 2 (1000 kWh) in the second decision period, again Li-ion battery should be installed at level 3 (3000 kWh) in the third period and finally vanadium redox battery should be installed at level 3 (3000 kWh). These results could be anticipated by looking at Tables 4 and 6. While the price difference of these two storage units is negligible in the initial phases, this difference becomes much more significant in the later stages. In these later stages, besides the lower price of vanadium redox battery, its excellent *DoD* plays a crucial role in making this type the preferred choice. These results are in accordance with the theoretical findings, which provide insights that vanadium redox battery could possibly hold special potential for future usage (IRENA, 2017). However, in scenario 2 there is a significant difference compared to scenario 1; investment in the last decision period goes to the lead-acid type, removing vanadium redox from the preferred choices. The reason behind this change is straightforward; the vanadium redox price did not decline as sharply as in scenario 1, making it therefore relatively expensive compared to cheaper options. In this situation, lead-acid became the dominant choice, despite its very low *DoD* value. Finally, in scenario 3 the dominance of vanadium redox battery type in all decision periods from 2 and on can be observed. During this scenario, all prices were declined sharply, except for the price of Li-ion battery type. Consequently, vanadium redox took its place and resulted in investments of Level 2, Level 3 and Level 2 again for the decision periods 2, 3, 4 respectively.

At this point, it would be useful to note two more conclusions that can be drawn from Fig. 6. Firstly, it is observed that there is a difference between the total installed capacity among the three scenarios; 7000 kWh in scenario 1 and 5000 kWh in scenarios 2, 3. While it would be expected that these values are equal, the results of Fig. 4a) should now be revised. There is a certain threshold after which installing more capacity of the same storage type does not result in significant savings. Therefore, that is why in scenario 2 the replacement of the vanadium redox installation at Level 3, was an installation of lead-acid at Level 2 instead of adding more capacity of Li-ion at Level 3. Of course, the same applies to scenario 3 and the case of vanadium redox battery. Lastly, it is also seen that there is no installation of flywheel energy storage system in any scenario. This happened because of the extremely high price of this specific storage type compared to its competitors. In the case study, where critical facilities are located in the microgrid and outages can last several hours, it is clear that someone can find more use in high energy density storage units. Flywheel storage systems can be considered as high-power and low-energy density units (Amiryar & Pullen, 2017). Of course, these results do not mean in any case that this specific storage type cannot find applications in the microgrid sector. Instead, they would be considered appropriate in situations where fast response is the top criterion for choosing storage options.

Finally, it would be worthwhile to compare and contrast the results of Fig. 6 with results of state-of-the-art models that can be found in the literature. Alsaidan et al. formulated a similar long-term expansion planning problem, specifically tailored for microgrid applications (Alsaidan et al., 2018). Although not all the assumptions are the same, the authors also considered four battery options in their case study, with two of them being also considered in the present work: Li-ion and lead-acid. In a high level, someone could say that their results are in accordance with the results presented here: in their case 1, there is a prevalence of Li-ion battery when project lifetime is longer and lead-acid option when it is shorter. It makes sense to assume that lead-acid battery would be more frequently present in the optimal policies of Fig. 6, if the problem horizon was decreased from 20 years to 10 years, given its low capital cost. However, the superiority of the proposed approach in this research is the fact that the policies we obtain are not static as opposed to (Alsaidan et al., 2018) or other similar models. Under different future conditions (i.e., disrupting technological advancements in the area of flow batteries, or a pause in the price decline of Li-ion batteries), the obtained optimal policies could be drastically different, as it can be shown in scenario 3 of Fig. 6. These are exactly the features that traditional fit-and-forget optimization approaches fail to encompass.

*4.3    Additional scenarios*

To elaborate more on the results obtained concerning optimal policies under various scenarios, it is considered suitable to analyze here a greater number of scenarios. These results are presented in the context of Fig. 7:

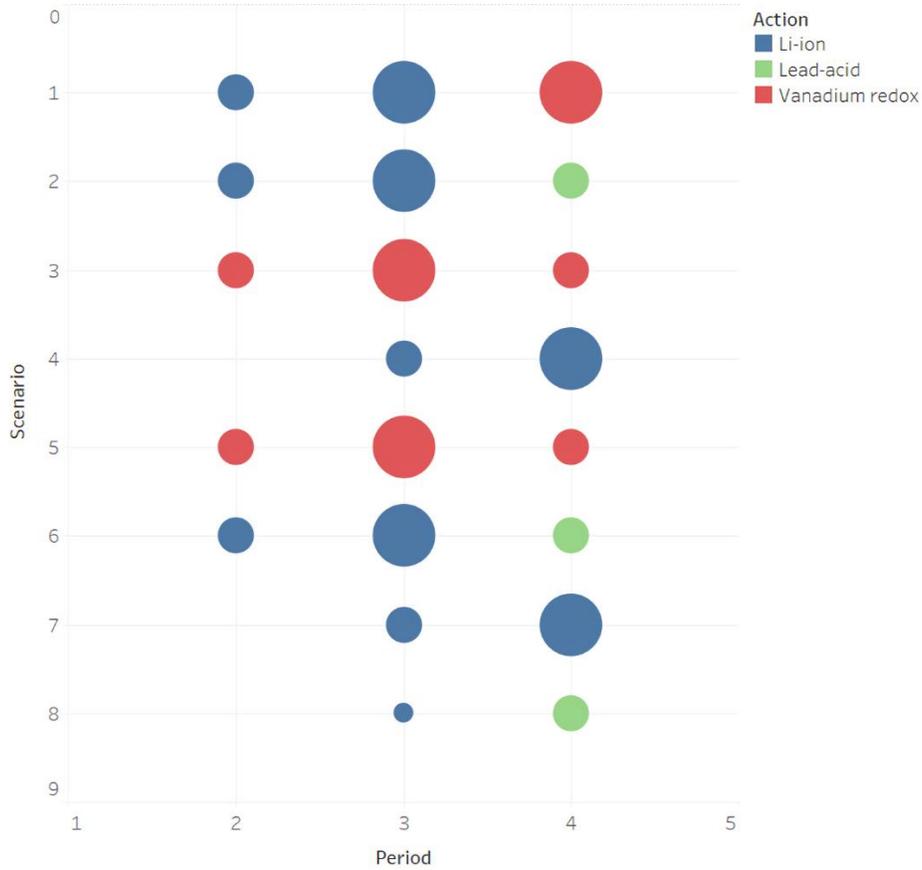

Fig. 7 Optimal policies for additional scenarios

It should be noted that the size of the circles in Fig. 7 denote the level at which at which the specific investment is made; small-sized circles correspond to Level 1 (300 kWh), moderate-sized circles correspond to Level 2 (1000 kWh) and large-sized circles correspond to Level 3 (3000 kWh). Fig. 7 contains a total of 8 scenarios: scenarios 1-3 correspond to the ones defined and studied before in Fig. 6. The rest of the scenarios in Fig. 7 correspond mainly to various combinations of price movements for the Li-ion and the vanadium redox battery. The reason for that is that the other two storage types examined are not able to become the dominant ones unless they gain a competitive advantage against the other two. More specifically, scenario 4 refers to the case where all the forward transitions are realized, except for Li-ion and vanadium redox batteries for periods 1 and 3. In scenario 5, Li-ion price remains constant throughout all periods, while all the other prices (except vanadium redox in period 2) are continuously declining. Moreover, in scenario 6, vanadium redox is the one that has its price remaining always constant, while the rest (except Li-ion in period 2) are again continuously declining. Scenario 7 corresponds to the case where all the prices are declining, but starting from period 2. Finally, scenario 8 has Li-ion and vanadium redox prices constant for periods 1, 2 while the rest have their prices constant only in period 1.

One important thing to notice here is that the only scenario in which the total installed capacity at the end of the time horizon is 7000 kWh is scenario 1, in which both Li-ion and vanadium redox batteries experience continuous declining trends. In all the other scenarios, the final obtained capacity was 5000 kWh, or even lower; for example, when the two dominant storage types' prices remained steady for the first two periods (scenario 8), the total installed capacity was way lower than typically. As mentioned before, flywheel energy storage was not chosen for any scenario. In another aspect, the high penetration of the vanadium redox battery depends heavily on its price movements; in the situations where this type of battery presented a steady behavior for at least two periods, the lead-acid battery was able to surpass it in the decision maker's choices even in cases where its own behavior remained steady for one period, like in scenario 8. Finally, it is obvious that the role of the Li-ion battery in energy systems such as the one examined in this case study is expected to remain crucial for the future. Nevertheless, there is a case where a potential level-off of Li-ion price, combined with a simultaneous decrease in vanadium redox price, could change the things in the hierarchy of these two storage types, as it happened in scenarios 3 and 5.

It has been made clear that proper uncertainty modeling is one of the main goals and originalities of the present work. To highlight the importance of uncertainty in the current problem, someone could examine again Fig. 6 and could compare specifically Scenario 2 and Scenario 3. It can be noticed that the only difference in these two scenarios is the price trajectory of vanadium redox/lithium-ion batteries in only two periods: 1,3. However, the optimization results are drastically different; there is not a single action that seems to agree with their coeval action from the other scenario.

### 4.4    *Evidence of convergence for Q-learning algorithm*

The final part of this section contains a crucially important check on whether the agent improves its experience with an increasing number of episodes. It should be reminded here that the number of episodes chosen for this experiment was $10^7$. However, the question here is how it can be asserted that this number of episodes was sufficient or not. Given the fact that the exploration/exploitation tradeoff parameter is decaying as a function of the number of episodes, it should always be expected to see improving performance of the agent as time passes by. However, the answer to this question can originate from running the experiment using a different number of episodes. The results of this procedure can be shown in Fig. 8:

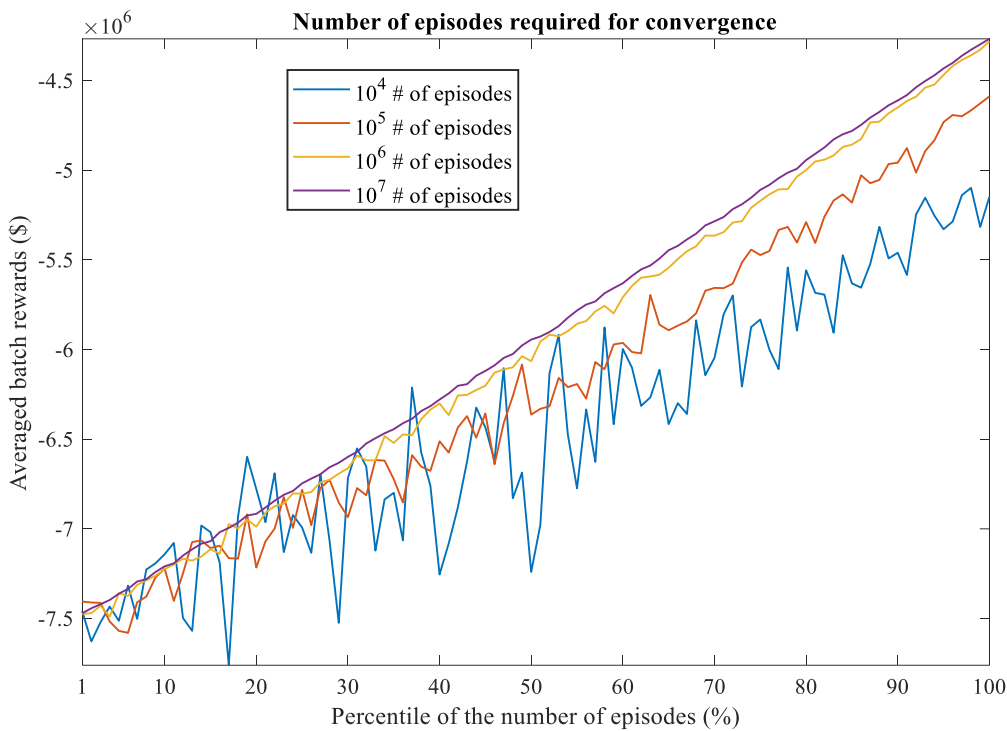

Fig. 8 Convergence check for the required number of episodes

In the horizontal axis of Fig. 8, the different percentiles of the total number of episodes are shown. In the vertical axis, the averaged total rewards can be seen for the corresponding batch of episodes belonging to that specific percentile. Of course, the exact number of episodes belonging to every percentile depends on the total number of episodes; for example, for the case of $10^6$ number of episodes, each batch contains a total of $10^4$ number of episodes. This is why different curves present different smoothness levels. However, Fig. 8 provides strong evidence that $10^7$ is a well-suited number for these research purposes. In order to see why specific focus should be given to the last 10 percentiles (90%-100%) in Fig. 8; this is exactly the region where the agent starts mostly to exploit its current knowledge and does not explore any more. In other words, the agent's performance becomes there as best as it can get. Therefore, it can be observed that the agent's performance is much worse in the case where $10^4$ or $10^5$ number of episodes were used. Now comparing the results for the situations of $10^6$ and $10^7$ number of episodes, someone could object that the difference is negligible. Nevertheless, given the scale of the problem, even this seemingly small margin constitutes a difference of several thousand dollars. In hypothetical scenarios where the dimensionality of the problem becomes even higher (which is true in more realistic settings), this difference could become even more significant. On the other hand, by using this logarithmic scale to compare different number of episodes required for convergence, it can definitely be assured that running these experiments for $10^8$ or more number of episodes would probably be a waste of computational resources.

*4.5    Applied value of the proposed model*

It is appropriate at this point to further elaborate and highlight the value of the proposed methodologies and approaches to a wide array of problems in the energy systems area.

Having well-defined frameworks for tackling a variety of long-term energy planning problems is of crucial importance. It is noted that such frameworks are currently missing from the literature and these are exactly the gaps that this work attempts to fill. For instance, except the storage expansion planning, the presented optimization approaches could be naturally extended to generation and transmission expansion planning. These two additional types of problems, although significantly different, are often studied together by energy researchers. The present work could be another decisive step towards facilitating the joint consideration of such optimization problems in the energy field. If this transition is successful, it could potentially result in significant savings for timing and monetary resources.

Overall, this work intends to serve as a baseline attempt to give a well-shaped structure to long-term planning problems which involve sequential decision-making investment. By doing so, the planners can have a clear glimpse at the future and therefore develop their investment strategy accordingly. However, this research can certainly inform policy making strategies as well. To further investigate this claim, the case study of the present work should be revisited; the research findings showed that vanadium redox flow battery could potentially play a significant role in the foreseen future of the energy storage, although that in order for this argument to be strengthened, further research is required towards the direction of relaxing as many simplistic assumptions as possible. As a consequence, governments may incentivize the businesses that are involved in the mining of vanadium by providing tax relief programs or any other means. Moreover, there may be incentives for microgrid planners as well, that intend to invest in vanadium redox storage units. To sum up, similar policy-making initiatives can be studied and recommended by leveraging the findings of the proposed framework in this paper as applied to the aforementioned classes of problems related to this work.

## 5. Conclusions

In this research, we develop a novel framework that is able to tackle sequential resource investment planning problems, such as the microgrid expansion planning. As a first step in this process, we formulate the problem as a Markov Decision Process, by properly defining states, actions and rewards. We solve this problem using a model-free Reinforcement Learning algorithm called Q-learning. We place particular emphasis on the overestimation bias issue of the Q-learning algorithm and deploy a simulation-based technique to mitigate this challenge. More specifically, we create a surrogate model that estimates the expected outage costs corresponding to specific states and actions of the environment. Finally, we conclude by giving important research findings, in terms of which storage technologies are expected to dominate the future of the energy storage area and also in terms of what the optimal capacity threshold is for a given microgrid setting.

### 5.1 Research contributions

The research contributions of the present work can be broadly divided into two distinct spaces. Firstly, from the energy systems perspective, we provide an original framework for the optimal storage sizing problem specifically tailored to renewable-based microgrid systems. In the wake of technological advancements which will consecutively bring lower storage investment costs, the significance of this research contribution will become even greater. Moreover, we formulate one of the first ever unified dynamic optimization problems which is able to derive optimal expansion policies for a finite time horizon. It is safe to assume that the analytical consideration and incorporation of stochastic modeling for several aspects of the problem are able to further illustrate the importance of this outcome. On the other hand, there are significant contributions from the operations management perspective, too. By providing analytical and detailed frameworks to tackle sequential resource investment planning problems, we can facilitate the whole pipeline in a wide array of problems in supply chain, operations strategy or resource allocation. Furthermore, from a more practical perspective, with the utilization of the approaches developed in the present work, it is possible to obtain more realistic and better solutions, from both an engineering and management perspective. This is due to the fact that the proposed approaches are able to capture a significantly greater amount of detail than the traditional ones. However, more research is required to fully evaluate the benefits of the information gained from this approach from the management side.

## 5.2    Research extensions

As the final step of this work, it is imperative to highlight a selection of potential extensions and improvements for the presented methodologies and frameworks. While the array of promising directions to explore is fairly wide, at this point the focus is given to two main ones: the exploration of several other fresh Reinforcement Learning-based techniques, as well as the inclusion of other and more diverse options for energy storage in the action set.

Concerning the former direction, it is clear that the given approach, although novel and promising, has its own set of limitations. The problem of overestimation bias is a good example and tackling it with simulation-based approaches is only one way. Moreover, it is true that complex and realistic planning agendas should be accompanied by equally complex and computationally efficient programming techniques. For all the above, there is recently an increasing attention to Deep Learning-based Reinforcement Learning algorithms, such as the double deep Q-learning. On top of that, there are also research attempts related to actor-critic methods in Reinforcement Learning, such as Deep Deterministic Policy Gradient (DDPG). Actor-critic methods are able to simultaneously estimate value functions (critic) and update accordingly the policy distribution (actor). Especially DDPG has the additive advantage that it handles continuous action spaces, which would be of course a great feature in the models. Finally, it is worthwhile mentioning Monte Carlo Tree Search (MCTS), another algorithm that is being studied and implemented recently mostly in game-based Reinforcement Learning. The basic idea of MCTS is to build a tree on all the possible scenarios of the simulation but explore only those that are the most promising ones. Overall, it should be mentioned that exploring new ideas, testing new algorithms and comparing them to ones that have been already implemented can only bring positive value to the subject.

Finally, concerning the latter extension, it is true that batteries are the dominant options whenever short-term energy storage is the main functionality desired. However, in order for a study in storage expansion planning to be complete from all possible angles, it should not omit to incorporate other energy storage types, as well. These alternative options include but are not limited to hydrogen storage, compressed air or pumped-storage hydroelectricity, with various advantages and disadvantages associated with each one of them. Although such options are not considered in the case study of the present work, it should be mentioned that the developed framework renders their consideration totally feasible in future works. In order to further strengthen this argument, it should be noted that this research plan should come accompanied by the algorithmic enhancements mentioned above. The result would be an enlargement in not only the action space of the problem, but also the state space with added components such as maintenance cost or efficiency decay. Conclusively, it is our plan to continue researching on new, modern, scalable and more efficient techniques which would also enable us to conduct case studies which are more and more realistic.